\pdfoutput=1

\documentclass[11pt]{article}

\usepackage[]{ACL2023}

\usepackage{times}
\usepackage{latexsym}

\usepackage[T1]{fontenc}

\usepackage[utf8]{inputenc}

\usepackage{microtype}

\usepackage{inconsolata}

\usepackage{soul}
\usepackage{graphicx}
\usepackage{hyperref}
\usepackage{amsmath}
\usepackage{ragged2e}
\usepackage{blindtext}
\usepackage{amsfonts}
\usepackage{multirow}
\usepackage{color}
\usepackage{booktabs, cellspace, multirow, tabularx}
\usepackage{array}
\usepackage{listings}
\usepackage[T1]{fontenc}
\usepackage{colortbl}
\usepackage{wrapfig,lipsum,booktabs}
\usepackage{amssymb}
\usepackage{pifont}
\usepackage{subfig}

\definecolor{airforceblue}{rgb}{0.36, 0.54, 0.66}
\definecolor{asparagus}{rgb}{0.94, 0.87, 0.8}
\definecolor{almond}{rgb}{0.53, 0.66, 0.42}
\definecolor{babyblue}{rgb}{0.54, 0.81, 0.94}
\definecolor{aureolin}{rgb}{0.99, 0.93, 0.0}

\definecolor{classname}{rgb}{0.98, 0.92, 0.84}
\definecolor{type}{rgb}{0.7, 0.75, 0.71}
\definecolor{synonym}{rgb}{0.96, 0.76, 0.76}
\definecolor{caption}{rgb}{0.54, 0.81, 0.94}
\definecolor{definition}{rgb}{0.67, 0.9, 0.93}
\definecolor{attribute}{rgb}{0.96, 0.73, 1.0}
\definecolor{track_path}{rgb}{1.0, 0.33, 0.64}

\newcommand{\cmark}{\ding{51}}%
\newcommand{\xmark}{\ding{55}}%
\title{Z-GMOT: Zero-shot Generic Multiple Object Tracking}

\author{
	Kim Hoang Tran$^{1,2}$, Anh Duy Le Dinh$^{1}$, Tien Phat Nguyen$^{1}$, Thinh Phan$^3$, \\ {\bf Pha Nguyen$^3$}, 
     {\bf Khoa Luu$^3$}, {\bf Donald Adjeroh$^4$}, {\bf Gianfranco Doretto$^4$}, {\bf Ngan Hoang Le$^3$} \\
	$^1$ FPT Software AI Center, Vietnam \\
         $^2$ VNUHCM-University of Science, Ho Chi Minh City, Vietnam \\
	$^3$ Department of Computer Science, University of Arkansas, USA\\
	$^4$ Department of Computer Science, West Virginia University, USA\\
}        

\begin{document}
\maketitle
\begin{abstract}
Despite recent significant progress, Multi-Object Tracking (MOT) faces limitations such as reliance on prior knowledge and predefined categories and struggles with unseen objects. To address these issues, Generic Multiple Object Tracking (GMOT) has emerged as an alternative approach, requiring less prior information. However, current GMOT methods often rely on initial bounding boxes and struggle to handle variations in factors such as viewpoint, lighting, occlusion, and scale, among others. 
Our contributions commence with the introduction of the \textit{Referring GMOT dataset} a collection of videos, each accompanied by detailed textual descriptions of their attributes. Subsequently, we propose $\mathtt{Z-GMOT}$, a cutting-edge tracking solution capable of tracking objects from \textit{never-seen categories} without the need of initial bounding boxes or predefined categories. Within our $\mathtt{Z-GMOT}$ framework, we introduce two novel components: (i) $\mathtt{iGLIP}$, an improved Grounded language-image pretraining, for accurately detecting unseen objects with specific characteristics. (ii) $\mathtt{MA-SORT}$, a novel object association approach that adeptly integrates motion and appearance-based matching strategies to tackle the complex task of tracking objects with high similarity. Our contributions are benchmarked through extensive experiments conducted on the Referring GMOT dataset for GMOT task. Additionally, to assess the generalizability of the proposed $\mathtt{Z-GMOT}$, we conduct ablation studies on the DanceTrack and MOT20 datasets for the MOT task. Our dataset, code, and models are released at: \url{https://fsoft-aic.github.io/Z-GMOT}.
\end{abstract}

\begin{figure}[thb]
\centering
\includegraphics[width=\columnwidth]{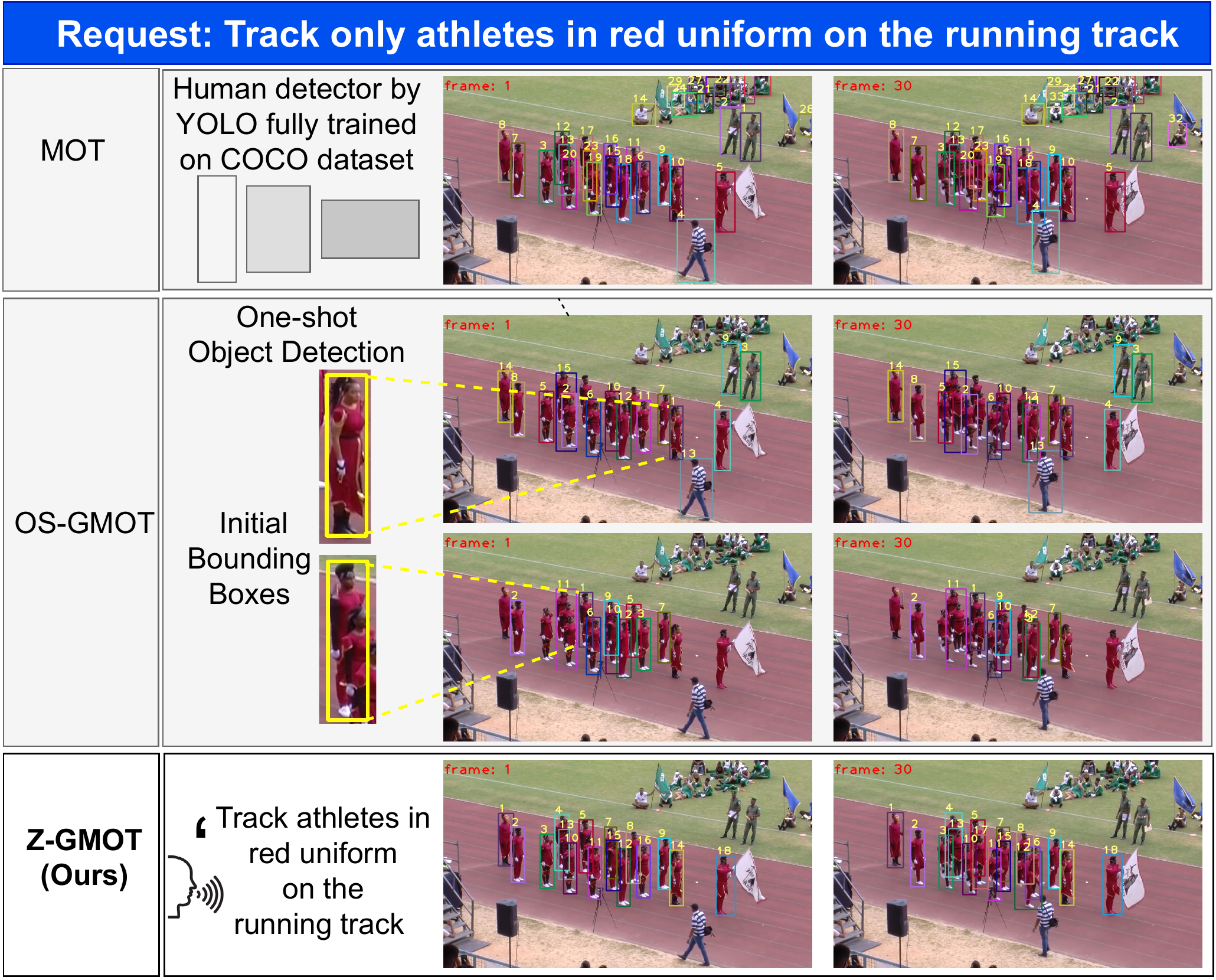}
\caption{High-level comparison between our $\mathtt{Z-GMOT}$ with conventional MOT and one-shot Generic MOT (OS-GMOT) for the task of tracking athletes in red uniforms on a running track. $1^{st}$ row: MOT, being a fully-supervised method, using YOLOX (trained on COCO) and OC-SORT (trained on DanceTrack) attempts to detect and track all people in the scene with high False Positive (FPs). $2^{nd}$ row: OS-GMOT is based on an initial bounding box and utilizes an MOT tracker (e.g. OS-SORT in this case). While reducing the number of FPs, OS-GMOT heavily relies on the initial bounding box, leading to variations in results with different bounding boxes and a high number of False Negatives (FNs). $3^{rd}$ row: our $\mathtt{Z-GMOT}$ including: (i) $\mathtt{iGLIP}$ effectively detects objects without the need for prior training or initial bounding boxes, and (ii) $\mathtt{MA-SORT}$ efficiently associates objects with high visual similarity.}
\label{fig:Teaser}
\end{figure}

\section{Introduction}
Multiple Object Tracking (MOT) \cite{bewley2016simple, leal2016learning, wojke2017simple, braso2020learning, wu2021track, cao2023observation, maggiolino2023deep, zhang2021bytetrack, yan2022towards, meinhardt2022trackformer, zeng2022motr, cai2022memot} aims to recognize, localize and track dynamic objects in a scene. It has become a cornerstone of dynamic scene analysis and is essential for many important real-world applications such as surveillance, security, autonomous driving, robotics, and biology. 


However, current MOT methods suffer from several limitations: they heavily depend on prior knowledge of tracking targets, requiring large labeled datasets; they struggle with tracking objects of unseen or specific categories; they are limited in handling objects with indistinguishable appearances. 
In contrast to MOT, Generic Multiple Object Tracking (GMOT) \cite{luo2013generic, luo2014bi} seeks to alleviate these challenges with reduced prior information. GMOT is tailored to track multiple objects of a shared or similar generic type, offering applicability in diverse domains like annotation, video editing, and animal behavior monitoring. Conventional GMOT methods \cite{luo2013generic, luo2014bi, bai2021gmot} adhere to a one-shot paradigm  \cite{huang2020globaltrack} and employ the initial bounding box of a single target object in the first frame to track all objects belonging to the same class. The conventional one-shot GMOT, which is based on one-shot object detection (OS-OD), is known as OS-GMOT. However, this approach heavily relies on the starting bounding box and has limitations in accommodating variations in object characteristics, including pose, illumination, occlusion, scale, texture, etc.

To overcome the aforementioned limitations of both MOT and OS-GMOT, particularly in the context of tracking multiple unseen objects without the requirement for training examples, we introduce a \textbf{novel tracking paradigm} called \textbf{Zero-shot Generic Multiple Object Tracking} ($\mathtt{Z-GMOT}$, which leverages recent advancements in Vision-Language (VL) models. Our $\mathtt{Z-GMOT}$ follows the tracking-by-detection paradigm and introduces two significant contributions aimed at enhancing both the object detection stage and object association stage. 

In the first stage, which involves object detection, we introduce an enhanced version of GLIP called $\mathtt{iGLIP}$. While GLIP has shown promise in detecting objects based on textual description queries, it faces limitations when tasked with detecting multiple objects with subtle distinguishing features. Specifically, our observations and empirical experiments have confirmed that it is sensitive to threshold settings, leading to high False Positives (FPs) at slightly lower thresholds and high False Negatives (FNs) at slightly higher thresholds. For instance, when asked to identify a ``red ball'' among multiple balls of various colors, GLIP may erroneously detect balls of different colors at a slightly lower threshold and miss the red ball when the threshold is increased only slightly. To address it, our proposed enhancement, $\mathtt{iGLIP}$, incorporates two distinct pathways. One pathway is tailored to handle general object categories like ``ball'', while the other pathway is dedicated to capturing specific object characteristics, such as the color ``red''. By integrating these dual pathways, $\mathtt{iGLIP}$ aims to deliver a more accurate and precise object detection process, especially when dealing with multiple generic objects.

In the second stage, which involves object association, we propose $\mathtt{{MA-SORT}}$ (\underline{M}otion-\underline{A}ppearance SORT), an innovative tracking algorithm that seamlessly fuses visual appearance with motion-based matching. $\mathtt{{MA-SORT}}$ adeptly measures appearance uniformity and dynamically balances the influence of motion and appearance during the association process.

Figure \ref{fig:Teaser} provides a visual comparison between our $\mathtt{Z-GMOT}$ with conventional MOT and OS-GMOT approaches with the task of tracking athletes in red uniforms on a running track as an example. In this comparison, MOT is a fully-supervised learning method that employs YOLOX object detection \cite{yolox2021} trained on COCO dataset \cite{chen2015microsoft} and OC-SORT object association \cite{cao2023observation} trained on DanceTrack dataset \cite{sun2022dancetrack}. Being a fully-supervised method, MOT attempts to detect and track all people in the scene instead of only athletes in red uniforms as requested. As a result, MOT generates a high number of FPs. In contrast, OS-GMOT relies on an initial bounding box to detect all requested objects. It also utilizes the robust OC-SORT tracker \cite{cao2023observation} for object association. While reducing the number of FPs, OS-GMOT heavily relies on the initial bounding box, leading to variations in results with different bounding boxes and a high number of False Negatives (FNs). Different from MOT and OS-GMOT, our $\mathtt{Z-GMOT}$ takes the tracking request in the form of a natural language description as its input to effectively detect and track objects without prior training or initial bounding boxes.
Our contributions are as follows:

\noindent
$\bullet$ We introduce a novel tracking paradigm $\mathtt{Z-GMOT}$, capable of tracking object categories that have never been seen before, all without the need for any training examples.

\noindent
$\bullet$ We present \textit{Referring GMOT dataset} consisting of \textit{Refer-GMOT40} and \textit{Refer-Animal} datasets. These datasets are built upon the foundations of the original GMOT-40 dataset \cite{bai2021gmot} and the AnimalTrack dataset \cite{zhang2022animaltrack} with the inclusion of natural language descriptions.

\noindent
$\bullet$ We propose $\mathtt{iGLIP}$ to effectively identifies unseen objects with specific characteristics.

\noindent
$\bullet$ We propose $\mathtt{MA-SORT}$, adeptly balancing between object motion and appearance to effectively track objects with highly similar appearances and complex motion patterns.

\noindent
$\bullet$ We conduct \textit{comprehensive experiments and ablation studies} on our newly introduced Referring GMOT dataset for GMOT task. We extend our experimentation to DanceTrack \cite{sun2022dancetrack}, MOT-20 \cite{dendorfer2020mot20} datasets for MOT tasks, to illustrate the effectiveness and generalizability of the proposed $\mathtt{Z-GMOT}$ framework.

\section{Related Works}
\label{sec:relate}
\noindent
\textbf{2.1 Pre-trained Vision-Language (VL) Models}
\vspace{0.5em}

Recent advancements in computer vision tasks have leveraged VL supervision, demonstrating remarkable transferability in enhancing model versatility and open-set recognition. A pioneering work in this domain is CLIP \cite{radford2021learning}, which effectively learns visual representations from vast amounts of raw image-text pairs. Since its release, CLIP has garnered significant attention \cite{yamazaki2022vlcap, yamazaki2023vltint, nguyen2023open, joo2023clip, yamazaki2023open, phan2024zeetad, 10446193, zhang2024vision}, and several other VL models, such as ALIGN \cite{jia2021scaling}, ViLD \cite{gu2022open}, RegionCLIP \cite{zhong2022regionclip}, GLIP \cite{li2022grounded, zhang2022glipv2}, Grounding DINO \cite{liu2023grounding}, UniCL \cite{yang2022unified}, X-DETR \cite{cai2022x}, OWL-ViT \cite{minderer2022simple}, LSeg \cite{li2022language}, DenseCLIP \cite{rao2022denseclip}, OpenSeg \cite{ghiasi2022open}, and MaskCLIP \cite{ding2022open}, have followed suit to signify a profound paradigm shift across various vision-related tasks. We can categorize VL pre-training models into three main groups: (i) Image classification: Models in this category, such as CLIP, ALIGN, and UniCL, are primarily focused on matching images with language descriptions through bidirectional supervised contrastive learning or one-to-one mappings. (ii) Object detection: This category encompasses models like ViLD, RegionCLIP, GLIPv2, X-DETR, and OWL-ViT, Grounding DINO, which tackle two sub-tasks: localization and recognition of objects within images. (iii) Image segmentation: The third group deals with pixel-level image classification by adapting pre-trained VL models, including models like LSeg, OpenSeg, and DenseSeg. \textit{In this work, we enhance GLIP and propose $\mathtt{iGLIP}$ to effectively capture object with specific characteristics.}

\vspace{0.5em}
\noindent
\textbf{2.2 Multiple Object Tracking (MOT)}
\vspace{0.5em}

Recent MOT approaches can be broadly categorized into two types based on whether object detection and association are performed by a single model or separate models, known respectively as joint detection and tracking and tracking-by-detection. In the first category \cite{CHAN2022108793, zhou2020tracking, pang2021quasi, wu2021track, yan2022towards, meinhardt2022trackformer, zeng2022motr, cai2022memot}, both objects detection and objects association are simultaneously produced in a single network. In this category, object detection can be modeled within a single network with re-ID feature extraction or motion features. In the second category \cite{bewley2016simple, leal2016learning, wojke2017simple, braso2020learning, cao2023observation, zhang2021bytetrack, nguyen2022multi, aharon2022bot, du2023strongsort, maggiolino2023deep, Cetintas_2023_CVPR}, an object detection algorithm performs detecting objects in a frame, then those objects are associated with previous frame tracklets to assign identities. It is important to note that the state-of-the-art (SOTA) in MOT has been dominated by the later paradigm. Our $\mathtt{Z-GMOT}$ approach falls under this paradigm. Particularly, we propose $\mathtt{iGLIP}$ for zero-shot objects detector and introduce $\mathtt{MA-SORT}$ for objects association generic objects with uniform appearances.

\vspace{0.5em}
\noindent
\textbf{2.3 Generic Multiple Object Tracking (GMOT)}
\vspace{0.5em}

In recent years, MOT has advanced significantly, but it remains tied to supervised learning prior knowledge and predefined categories, complicating the tracking of unfamiliar objects. Different from MOT, GMOT \cite{luo2013generic, luo2014bi, bai2021gmot} aims to alleviate MOT's limitations by reducing the dependency on prior information. GMOT is designed to track multiple objects of a common or similar generic type, making it suitable for a wide array of applications, ranging from annotation and video editing to monitoring animal behavior. Thus, GMOT often deals with scenarios where objects appear in groups (such as a herd of cows, a school of fish, or a swarm of ants). Consequently, GMOT faces various challenges, including dense object scenarios, small objects, objects with occlusions, among other complexities. Notwithstanding, conventional GMOT methodologies \cite{luo2013generic, luo2014bi, bai2021gmot} are predominantly anchored in a one-shot paradigm, i.e. OS-GMOT, leveraging the initial bounding box of a single target object in the first frame to track all objects of the same class. While OS-GMOT shows promise by requiring less prior information, it heavily relies on initial bounding boxes and struggles with viewpoint, lighting, occlusion, and scale variations. Different from MOT (fully-supervised) and OS-GMOT (using initial bounding box), we \textit{introduce a novel zero-shot tracking paradigm known as $\mathtt{Z-GMOT}$. Our $\mathtt{Z-GMOT}$ enables users to track multiple generic objects in videos using natural language descriptors, without the need for prior training data or predefined categories.}

\section{Referring GMOT dataset}
\label{sec:dataset}

\begin{table}[!thb]
\caption{Comparison of \textbf{existing datasets} of SOT, MOT, GSOT, GMOT. ``\#" represents the quantity of the respective items. Cat., Vid. denote Categories and Videos. NLP indicates textual natural language descriptions.}
\vspace{-1em}
\setlength{\tabcolsep}{1pt}
\renewcommand{\arraystretch}{1.1}
\resizebox{\linewidth}{!}{
\begin{tabular}{c|l|l||lllll}
\toprule
& \textbf{Datasets}   & \textbf{NLP} & \#\textbf{Cat.} & \#\textbf{Vid.} & \#\textbf{Frames} & \#\textbf{Tracks} & \#\textbf{Boxs} \\ \midrule
\rowcolor{asparagus!30}
 & OTB2013~\cite{wu2013online}  &  \xmark   &       10       &    51      &      29K    &      51    &    29K   \\ 
\rowcolor{asparagus!30}
& VOT2017~\cite{kristan2016novel}   &  \xmark   &      24         &    60      &     21K      &     60     &   21K    \\ 
\rowcolor{asparagus!30}
SOT & TrackingNet~\cite{muller2018trackingnet}  &  \xmark   &       21       &    31K      &   14M       &   31K       &    14M   \\ \cline{2-8} 
\rowcolor{asparagus!80}
& LaSOT~\cite{fan2019lasot}      &  \cmark    &      70        &    1.4K      &     3.52M     &    1.4K      &     3.52M  \\ 
\rowcolor{asparagus!80}
& TNL2K~\cite{wang2021towards}     & \cmark    &         -     &    2K      &      1.24M    &    2K      &   1.24M    \\ \midrule \midrule
\rowcolor{babyblue!20}
& 
MOT17~\cite{milan2016mot16}    &  \xmark   &      1        &    14      &    11.2K      &    1.3K      &   0.3M    \\ 
\rowcolor{babyblue!20}
& MOT20~\cite{dendorfer2020mot20}    &  \xmark    &        1      &     8     &     13.41K     &     3.45K     &     1.65M  \\ 
\rowcolor{babyblue!20}
& Omni-MOT~\cite{sun2020simultaneous}    &  \xmark    &          1    &     -     &      14M+    &    250K      &    110M   \\ 
\rowcolor{babyblue!20}
MOT & DanceTrack~\cite{sun2022dancetrack}  &  \xmark    &     1         &    100      &     105K     &    990      &   -    \\  
\rowcolor{babyblue!20}
& TAO~\cite{dave2020tao}   &  \xmark    &       833       &     2.9K     &    2.6M      &    17.2K      &    333K   \\
\rowcolor{babyblue!20}
& SportMOT~\cite{cui2023sportsmot}    &  \xmark    &       1       &    240      &    150K      &     3.4K     &    1.62M   \\  \cline{2-8}
\rowcolor{babyblue!70}
& Refer-KITTI~\cite{wu2023referring}   & \cmark    &     2         &     18     &    6.65K      &   637       &   28.72K    \\ \midrule \midrule
\rowcolor{almond!30}
GSOT& 
GOT-10~\cite{huang2019got}    &  \xmark    &      563        &    10K      &    1.5M      &     10K      &    1.5M   \\ 
\rowcolor{almond!30}
& Fish~\cite{kay2022caltech}    &  \xmark    &      1        &   1.6K       &  527.2K        &     8.25k      &    516K   \\ \midrule \midrule
\rowcolor{aureolin!20} & 
AnimalTrack~\cite{zhang2022animaltrack}  &  \xmark   &      10        &     58     &   24.7K       &   1.92K       &    429K   \\ 
\rowcolor{aureolin!20}
GMOT& GMOT-40~\cite{bai2021gmot}  &  \xmark   &        10      &    40      &    9K      &   2.02K       &  256K     \\ \cline{2-8}
\rowcolor{aureolin!50}
& \textbf{Refer-Animal(Ours)} &  \cmark   &      10        &     58     &   24.7K       &   1.92K       &    429K   \\ 
\rowcolor{aureolin!50}
& \textbf{Refer-GMOT40(Ours)}  &  \cmark   &        10      &    40      &    9K      &   2.02K       &  256K     \\
\bottomrule
\end{tabular}}
\vspace{-1em}
\label{tb:dataset_comparison}
\end{table}

Table \ref{tb:dataset_comparison} presents statistical information for existing tracking datasets including Single Object Tracking (SOT), Generic Single Object Tracking (GSOT), MOT, GMOT. With the recent advancements and the capabilities of Large Language Models (LLMs), there's a growing demand for including textual descriptions in tracking datasets. While natural language have already found their place in SOT and MOT datasets, they have been conspicuously absent from GMOT datasets until now. As a result, our dataset is the pioneering effort to address this demand, integrating textual descriptions into the GMOT domain for the first time.

\begin{figure}[!thb]
\centering
\includegraphics[width=\columnwidth]{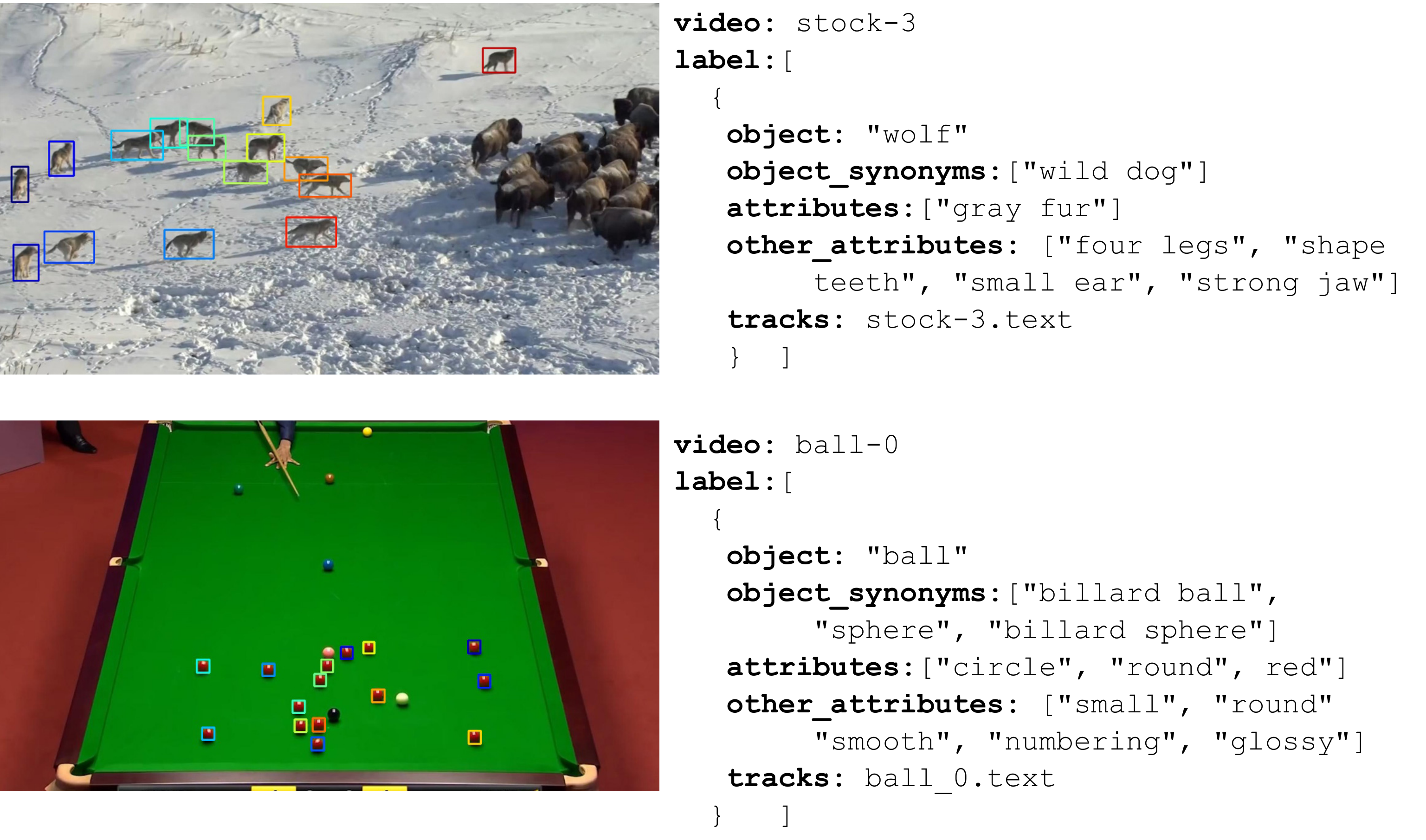}
\vspace{-2em}
\caption{Examples of data annotation structure. \vspace{-1em}}

\label{fig:data_example}
\end{figure}

In this work, we propose to incorporate textual descriptions into two pre-existing GMOT datasets, namely GMOT-40 \cite{bai2021gmot} and AnimalTrack \cite{zhang2022animaltrack}, and designate them as the \textit{``Refer-GMOT40''} and \textit{``Refer-Animal''} datasets. \textit{Refer-GMOT40} consists of 40 videos featuring 10 real-world object categories, each containing 4 sequences. \textit{Refer-Animal} contains 26 video sequences depicting 10 prevalent animal categories. Each video undergoes annotation, comprising of an \texttt{\textbf{object}} name, its corresponding \texttt{\textbf{attributes}} description, and its corresponding \texttt{\textbf{tracks}}. It's worth emphasizing that the \texttt{\textbf{attributes}} description primarily focuses on discernible object characteristics, while \texttt{\textbf{other\_attributes}} aims to offer additional details about the object's traits. Importantly, some of the attributes listed under \texttt{\textbf{other\_attributes}} may not always be visible throughout the entirety of the video.
To maintain the standardized format for MOT challenges, as outlined in \cite{milan2016mot16, dendorfer2020mot20}, each video comes with its tracking ground truth, stored in a separate text file within \texttt{\textbf{tracks}} annotation. This approach ensures consistency with MOT problem conventions. The annotation process follows the JSON format, and Figure \ref{fig:data_example} offers illustrative examples of the annotation structure. This data is conducted by 4 annotators and made publicly available. 

\section{Proposed $\mathtt{Z-GMOT}$}
\label{sec:proposed}
Our $\mathtt{Z-GMOT}$ framework follows the tracking-by-detection paradigm which includes the object detection stage and object association one. In the initial stage, we analyze the limitations of GLIP detector which is our motivation for proposing $\mathtt{iGLIP}$ for detecting effectively generic objects. In the subsequent stage, we introduce $\mathtt{MA-SORT}$ to adeptly balance between motion cues and visual appearances to improve the association process.

\subsection{Proposed $\mathtt{iGLIP}$}
We start by analyzing the limitations of GLIP and then proposing $\mathtt{iGLIP}$.

\begin{figure}[thb]
\centering
\includegraphics[width=1.0\columnwidth]{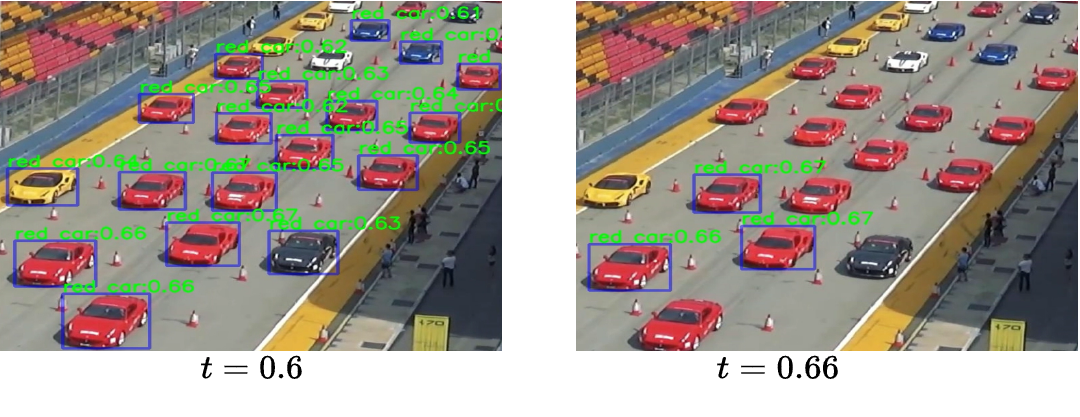}
\vspace{-2em}
\caption{Limitation 1 of GLIP: Sensitive to threshold selection. With slightly different thresholds $t=0.6$ v.s. $t=0.66$, GLIP produces different results with high FPs (left) and high FNs (right). Note that GLIP uses prompt \emph{``red car''}) in both results.}
\label{fig:limitations_1}
\end{figure}

\begin{figure}[thb]
\centering
\includegraphics[width=1.0\columnwidth]{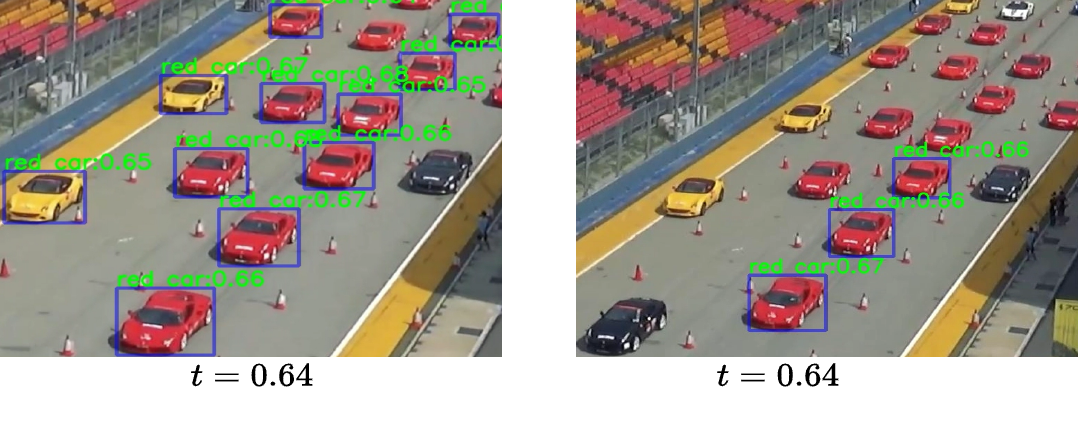}
\vspace{-2em}
\caption{Limitation 2 of GLIP: With the same $t$. ($t = 0.64$) and the same prompt \emph{``red car''}, the results vary when applied to two similar input images.}
\label{fig:limitations_2}
\end{figure}

\noindent
\textbf{Limitations of GLIP.}
GLIP encounters difficulties in handling specific object categories ($OC^{Spe}$) characterized by attributes. As shown in Fig. \ref{fig:limitations_1}, object detection performance displays sensitivity to threshold selection; even slight threshold variations lead to significant outcome differences. This leads to high TPs at slightly lower threshold and high FNs at slightly higher threshold. Fig. \ref{fig:limitations_2} underscores GLIP's drawbacks in effectively capturing objects with specific attributes. Even when using the same threshold selection and prompt, the outcomes exhibit variations on similar images with specific object category $OC^{Spe}$.

\begin{figure*}[t]
\centering
\includegraphics[width=1.0\textwidth]{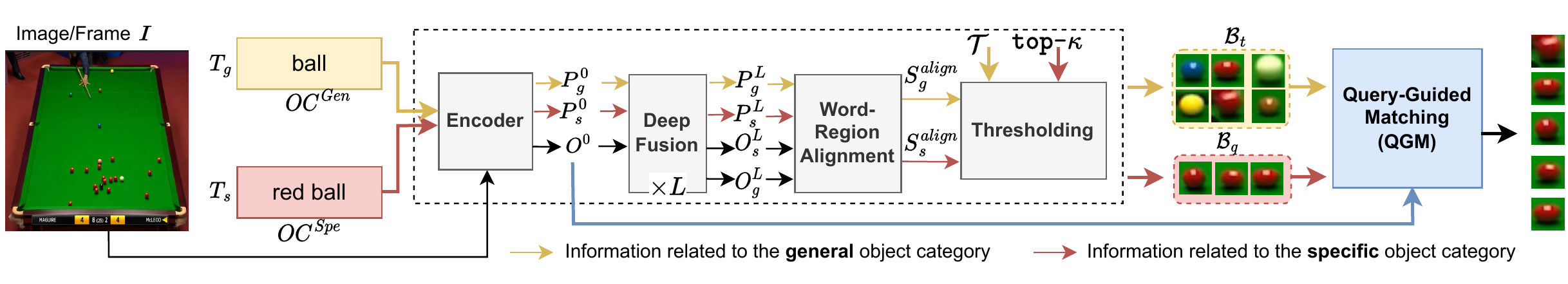}
\vspace{-2em}
\caption{Network architecture of $\mathtt{iGLIP}$, which inputs an image $I$, a general prompt $T_g$ (e.g. ``ball''), and a specific prompt $T_s$ (e.g. ``red ball''). $\mathtt{iGLIP}$ includes a QGM module to eliminate FPs generated from the general prompt.}
\label{fig:GLIP_iGLIP}
\end{figure*}

\begin{figure}[t]
\centering
\includegraphics[width=1.0\columnwidth]{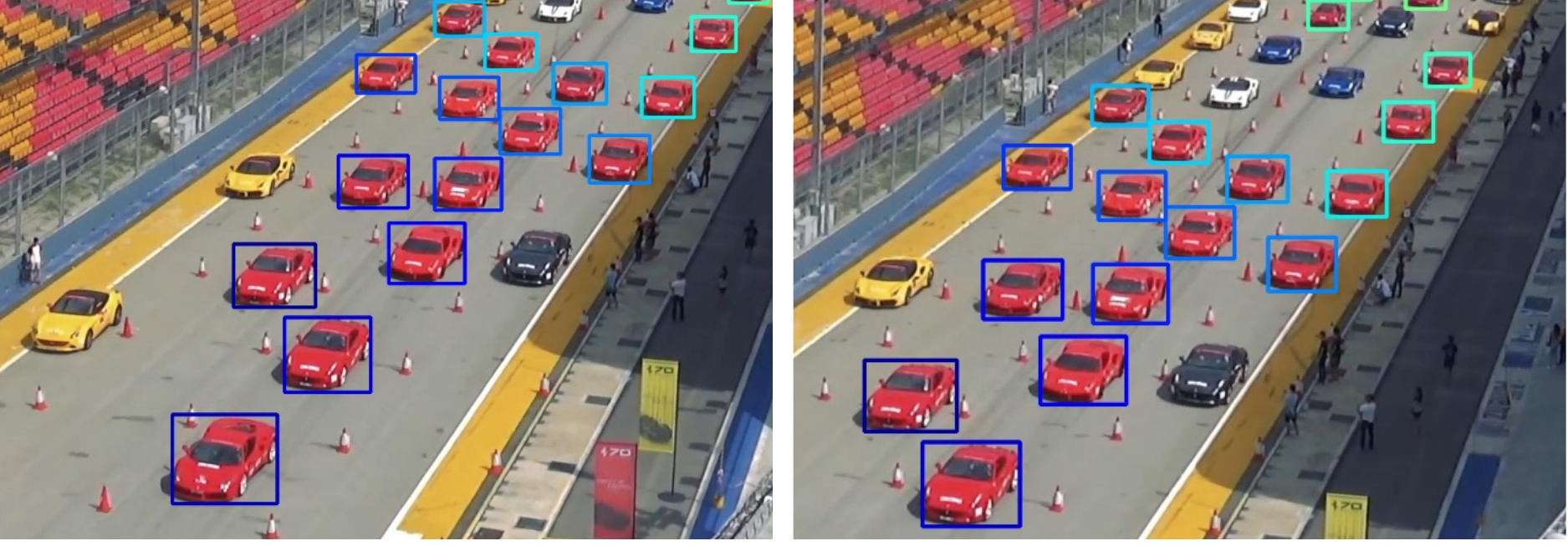}
\caption{Detection by $\mathtt{iGLIP}$ with general prompt $T_g$ as ``car'' and specific prompt $T_s$ as ``red car'' across images.}
\label{fig:iGLIP-redcar}
\end{figure}

\noindent
\textbf{Proposed $\mathtt{iGLIP}$}.
As depicted in Figure \ref{fig:GLIP_iGLIP}, our proposed $\mathtt{iGLIP}$ takes an input image $I$ and two kinds of prompt, namely, a specific prompt ($T_s$) for $OC^{Spe}$ and a general prompt ($T_g$) for $OC^{Gen}$. Both $T_s$ and $T_g$ are derived from the Refering GMOT dataset. Herein, $T_g$ is set as the \texttt{\textbf{object}} (e.g.,``ball''), while $T_s$ is defined as a combination of \texttt{\textbf{attributes}} and \texttt{\textbf{object}} (e.g., ``red ball'').
Both prompts $T_s$ and $T_g$ go through a text encoder, i.e., BERTModule \cite{devlin2018bert} to obtain contextual word features $P_s^0$ and $P_g^0$, respectively. Meanwhile, the image goes through a visual encoder, i.e., Swin \cite{liu2021swin} to obtain proposal features $O^0$. Then, $L$ deep fusion layers \cite{li2022grounded} are applied into contextual word features $P_s^0$, $P_g^0$ and $O^0$. The $i^{th}$ layer of deep fusion is as follows: 
\begin{subequations}
\begin{align}
{O^i_{s-\text{t2i}}, P^i_{s-\text{i2t}} } = \text{X-MHA}&(O_s^i, P_s^i)\\
{ O^i_{g-\text{t2i}}, P^i_{g-\text{i2t}} } = \text{X-MHA}&(O_g^i, P_g^i), 
\end{align}
\end{subequations}
, where specific proposal features are represented by
$O_s^{i+1} = \text{DyHeadModule}(O_s^i { + O^i_{s-\text{t2i}}})$, general proposal features are denoted by
$O_g^{i+1} = \text{DyHeadModule}(O_g^i { + O^i_{g-\text{t2i}}})$, and specific contextual word features
$P_s^{i+1} = \text{BERTModule}(P_s^i { + P^i_{s-\text{i2t}}})$, general contextual word features $P_g^{i+1}  =
\text{BERTModule}(P_g^i { + P^i_{g-\text{i2t}}})$. Notably, where $L$ is the number of DyHeadModules in DyHead \cite{Dai_2021_CVPR} and $O_s^0 = O_g^0 = O^0$. X-MHA denotes a cross-modality multi-head attention module. Finally, the word-region alignment module is utilized to compute the alignment score using dot product between the fused features. 
\begin{align}\label{eqn:ground_logits}
    & S_s^{\text{align}} = O_s P_s^{\top}, \text{ and } 
    & S_g^{\text{align}} = O_g P_g^{\top}
\end{align}
where $O_s = O_s^L \in \mathbb{R}^{N \times d}, O_g = O_g^L \in \mathbb{R}^{N \times d}$ are the visual features from the last visual encoder layer and $P_s = P_s^L\in \mathbb{R}^{M \times d}, P_g = P_g^L \in \mathbb{R}^{M \times d}$ are the word features of $OC^{Spe}$ and $OC^{Gen}$ from the last language encoder layer. The result of this operation are matrices $S_s^{\text{align}} \in \mathbb{R}^{N \times M}, S_g^{\text{align}} \in \mathbb{R}^{N \times M}$. 
The resulting bounding boxes undergo a filtering process using two parameters: top-$\mathtt{\kappa}$ and threshold $\mathcal{T}$. The top-$\mathtt{\kappa}$ parameter is applied into $S_s^{\text{align}}$ to extract a set of queries $\mathcal{B}_q$, which represents template patterns. In order to exclusively detect TPs, we have set $\mathtt{\kappa} = 5$. The threshold $\mathcal{T}$ parameter is applied into $S_g^{\text{align}}$ to extract a target set $\mathcal{B}_t$. To capture all object proposals, even those potentially including FPs, we set $\mathcal{T} = 0.3$.
Query-Guided Matching ({QGM}) module is then proposed to eliminate FPs in $\mathcal{B}_t$ by using $\mathcal{B}_q$ as template patterns. To perform QGM matching without adding additional cost, we propose to utilize only visual features $O^0$ extracted from the backbone, without the influence of text embeddings, to ensure the feature is enriched with visual properties. Let $O^0_t$ and $O^0_q$ represent the visual features of object proposals in $\mathcal{B}_t$ and $\mathcal{B}_q$, the matching score is defined as the cosine similarity:
\begin{equation}
    S_{qt} \!=\! \cos{(O^0_q \cdot {O^0_t}^T)}.
\label{eq:simm}
\end{equation}
The final detection results comprise the query objects and candidate objects with high similarity. Figure \ref{fig:iGLIP-redcar} illustrates the red car detection by our proposed $\mathtt{iGLIP}$ with general prompt $T_g$ as ``car'' and specific prompt $T_s$ as ``red car'' across all images. All results in Figure \ref{fig:iGLIP-redcar} are generated with the same default settings of $\mathcal{T} = 0.3$ and $\mathtt{\kappa} = 5$.

\subsection{Proposed $\mathtt{MA-SORT}$}
In this section, we introduce our proposed tracking method - 
\textit{MA-SORT: Balance visual appearance and motion cues: }The standard similarity between $N$ existing track and $M$ detected box embeddings is defined using cosine distance, $C_a \in \mathbb{R}^{M\times N}$. In a typical tracking approach that combines visual appearance and motion cues, the cost matrix $C$ is computed as $C = M_c + \alpha C_a$, where $M_c$ represents the motion cost, measured by the IoU cost matrix. Leveraging DeepOC-SORT \cite{maggiolino2023deep}, which computes a virtual trajectory over the occlusion period to rectify the error accumulation of filter parameters during occlusions, the matrix cost becomes:
\begin{equation}
C = IoU + \lambda C_v + \alpha C_a,
\label{eq:cost_0}
\end{equation}
where $C_v$ represents the consistency between the directions of i) linking two observations on an existing track, and ii) linking tracks' historical observations and new observations. $\lambda$ and $\alpha$ are hyperparameters to determine the significance of motion and visual appearance, respectively.


To strike a balance between visual appearance and motion cues, we incorporate appearance weight $W_{a}$ and motion weight $W_{m}$ into Eq.\ref{eq:cost_0}. To effectively handle the high similarity between objects of the same generic type in GMOT, we propose the following hypothesis: when the visual appearances of all detections are very similar, the tracker should prioritize motion over appearance. The homogeneity of visual appearances across all detections can be quantified as follows:
\begin{equation}
\mu = \frac{1}{M}\sum_{i=1}^{M}{f_i} \text{ and }
\mu_{det} = \frac{1}{M}\sum_{i=1}^{M}{\mathtt{cos}(f_i, \mu)}.
\label{eq:distribution}
\end{equation}
Where $M$ is the number of detections in a frame, $f_i$ is a feature vector of the $i$-th detection gained from re-ID model \cite{Wojke2018deep}.

Here, we consider $\theta$ as a vector distance threshold to determine the similarity between two vectors; if the angle between them is smaller than $\theta$, the vectors are considered more similar.
\begin{equation}
    W_{a} = \frac{(1 - \mu_{det})}{1-\mathtt{cos}(\theta)}.
\label{eq:theta}
\end{equation}
We initialize $W_{m}$ as 1, indicating that both motion and appearance are equally important. As $W_{a}$ decreases, we propose redistributing the remaining weight to motion, $W_{m}$: 
\begin{equation}
    W_{m} = 1 + \left[1 - W_{a}\right] = 2 - \frac{(1  -  \mu_{det})}{1-\mathtt{cos}(\theta)}.
\end{equation}

As a result, the final cost matrix $C$ is:
\begin{equation}
\begin{aligned}
C = W_{m}(IoU + \lambda C_v) + W_{a}C_a.
\end{aligned}
\label{eq:cost_1}
\end{equation}

\section{Experimental Results}
\label{sec:experiment}
\subsection{Datasets, Metrics and Experiment Details}

We assess our $\mathtt{Z-GMOT}$ framework on our Referring GMOT dataset for the GMOT task. To demonstrate the generalizability of $\mathtt{Z-GMOT}$ framework, we extend our evaluation to include \textit{DanceTrack} \cite{sun2022dancetrack} and \textit{MOT20} \cite{dendorfer2020mot20} for the MOT task. \textit{Refering GMOT dataset}, consisting of \textit{Refer-GMOT40} and \textit{Refer-Animal} dataset, is described in Section \ref{sec:dataset}. \textit{DanceTrack} is a vast dataset designed for multi-human tracking i.e., group dancing. It includes 40 train, 24 validation, and 35 test videos, totaling 105,855 frames recorded at 20 FPS. \textit{MOT20} is an updated version of MOT17 \cite{milan2016mot16} including more crowded scenes, object occlusion, and smaller object size than MOT17.

We employ the following metrics: Higher Order Tracking Accuracy ($HOTA$) \cite{Luiten_2020hota}, Multiple Object Tracking Accuracy ($MOTA$) \cite{Bernardin2008mota}, and $IDF1$ \cite{Ristani2016idf1}. $HOTA$ is measured based on Detection Accuracy ($DetA$), Association Accuracy ($AssA$), i.e. $HOTA = \sqrt{DetA \cdot AssA}$, thus, it effectively strikes a balance in assessing both frame-level detection and temporal association performance.
All experiments and comparisons have been conducted by an NVIDIA A100-SXM4-80GB GPU.

\begin{table}[!t]
\setlength{\tabcolsep}{4pt}
\renewcommand{\arraystretch}{1.2}
  \centering
  \caption{\underline{Tracking comparison} on \textit{Refer-GMOT40} dataset between our $\mathtt{iGLIP}$ with SOTA OS-OD \cite{bai2021gmot} on various trackers. For each tracker, the best scores are highlighted in \textbf{bold}.}
  \vspace{-1em}
  \label{tab:gmot40}
  \resizebox{\columnwidth}{!}{%
\begin{tabular}{l|l|c|ccc}
\toprule
\textbf{Trackers}   & \textbf{Detectors} & \textbf{\#-Shot} & \textbf{HOTA}$\uparrow$ & \textbf{MOTA}$\uparrow$ & \textbf{IDF1}$\uparrow$ \\ \hline
SORT  &   OS-OD &    one-shot   &       30.05          & 20.83       &      33.90    \\ 
\cite{bewley2016simple} & \textbf{$\mathtt{iGLIP}$}\textbf{(Ours)} & zero-shot & \textbf{54.21} & \textbf{62.90} & \textbf{64.34}  \\ \hline

{DeepSORT}       & OS-OD  & one-shot & 27.82          & 17.96    & 30.37\\   
\cite{wojke2017simple}    & $\mathtt{iGLIP}$\textbf{(Ours)} & zero-shot &  \textbf{50.45} &  \textbf{58.99}  & \textbf{57.55} \\ \hline

{ByteTrack}     & OS-OD   & one-shot  & 29.89          & 20.30    & 34.70  \\   
\cite{zhang2021bytetrack}    & $\mathtt{iGLIP}$\textbf{(Ours)}& zero-shot & \textbf{53.69} & \textbf{61.49} & \textbf{66.21}\\ \hline

{OC-SORT}        & OS-OD  & one-shot & 30.35 & 20.60    & 34.37 \\   
\cite{cao2023observation}    & $\mathtt{iGLIP}$\textbf{(Ours)} &  zero-shot &  \textbf{56.51} & \textbf{62.76} & \textbf{67.40} \\ \hline
    
{\shortstack{Deep-OCSORT}}       & OS-OD   & one-shot & 30.37 & 21.10 & 35.12  \\   
\cite{maggiolino2023deep}& $\mathtt{iGLIP}$\textbf{(Ours)}&  zero-shot &  \textbf{55.89} & \textbf{64.02} & \textbf{66.52} \\ \hline

{MOTRv2}        & OS-OD  & one-shot & 23.75 & 13.87  & 25.17   \\   
\cite{zhang2023motrv2}    & $\mathtt{iGLIP}$\textbf{(Ours)} &  zero-shot  & \textbf{31.32}  & \textbf{18.54} & \textbf{31.28}  \\
\bottomrule
\end{tabular}}
\label{tb:GMOT40-iGLIP}
\end{table}

\begin{table}[!t]
\setlength{\tabcolsep}{5pt}
\renewcommand{\arraystretch}{1.2}
  \centering
  \caption{\underline{Tracking comparison} on \textit{Refer-GMOT40} dataset between our $\mathtt{MA-SORT}$ with other trackers. Our proposed $\mathtt{iGLIP}$ is used as the object detection. The best scores are highlighted in \textbf{bold}.}
  \vspace{-1em}
  \label{tab:gmot40}
  \resizebox{\columnwidth}{!}{%
\begin{tabular}{l|ccc}
\toprule
\textbf{Trackers}  &  \textbf{HOTA}$\uparrow$ & \textbf{MOTA}$\uparrow$ & \textbf{IDF1}$\uparrow$ \\ \hline
SORT \cite{bewley2016simple} & {54.21} & {62.90} & {64.34}  \\  

DeepSORT \cite{wojke2017simple} &  {50.45} &  {58.99}  & {57.55}  \\   
ByteTrack \cite{zhang2021bytetrack}   & {53.69} & {61.49} & {66.21} \\  
OC-SORT \cite{cao2023observation}   & {56.51} & {62.76} & {67.40}  \\   
Deep-OCSORT \cite{maggiolino2023deep} &  {55.89} & {64.02} & {66.52} \\  
MOTRv2 \cite{zhang2023motrv2} & 31.32  & 18.54 & 31.28 \\ \hline
$\mathtt{MA-SORT}$\textbf{(Ours)} & \textbf{56.75} & \textbf{64.62} & \textbf{68.17}  \\
\bottomrule
\end{tabular}}
\label{tb:GMOT40-MASORT}
\end{table}


\begin{table}[!t]
\setlength{\tabcolsep}{2pt}
\renewcommand{\arraystretch}{1.2}
  \centering
  \caption{\underline{Tracking comparison} on \textit{Refer-Animal} between our $\mathtt{Z-GMOT}$ and existing \textit{fully-supervised MOT} methods. The best scores are highlighted in \textbf{bold}.}
  \vspace{-1em}
  \resizebox{1.0\linewidth}{!}{
    \begin{tabular}{l|l|c|ccc}
    \toprule
    \textbf{Tracker}                    & \textbf{Detector}    & \textbf{Train}                 & \textbf{HOTA}$\uparrow$                  & \textbf{MOTA}$\uparrow$                  & \textbf{IDF1}$\uparrow$                             \\ \hline

    \multirow{1}{*}{SORT}      & FRCNN\cite{ren2015faster} & \cmark & 42.80                           & 55.60                           & 49.20                   \\
    \multirow{1}{*}{DeepSORT}  & FRCNN\cite{ren2015faster} & \cmark & 32.80                           & 41.40                           & 35.20                                         \\
    \multirow{1}{*}{ByteTrack} & YOLOX\cite{yolox2021}       & \cmark & 40.10                           & 38.50                           & 51.20                                                   \\ 
    \multirow{1}{*}{TransTrack} & YOLOX\cite{yolox2021}       & \cmark & 45.40                           & 48.30                           & 53.40                                                   \\
    \multirow{1}{*}{QDTrack} & YOLOX\cite{yolox2021}       & \cmark & 47.00                           & 55.70                           & 56.30                                                   \\    
    \hline

    $\mathtt{MA-SORT}$\textbf{(Ours)}&  YOLOX\cite{yolox2021}      & \cmark &         \textbf{57.86}                   &      \textbf{68.32}                    &  \textbf{63.01}        \\    
    $\mathtt{MA-SORT}$\textbf{(Ours)}&  $\mathtt{iGLIP}$ $\mathtt{(Z-GMOT)}$\textbf{(Ours)}      & \xmark &        53.28                    &           57.64                 &       58.43  \\ \bottomrule              
    \end{tabular}
}
  \label{tb:Animal-ZGMOT}%
\end{table}%

\begin{table}[!ht]
\centering
\setlength{\tabcolsep}{1pt}
\renewcommand{\arraystretch}{1.2}
\caption{\underline{Ablation study of generalizability} of $\mathtt{Z-GMOT}$ on \textit{DanceTrack} validation set with \textit{MOT tas}k.}
\vspace{-1em}
    \resizebox{1.0\linewidth}{!}{\begin{tabular}{l|l|c|cccc}
    \hline
    {\textbf{Trackers}}   & {\textbf{Detectors}} & {\textbf{Train}} &  \textbf{HOTA}$\uparrow$   & \textbf{MOTA}$\uparrow$      & \textbf{IDF1}$\uparrow$      \\ \hline
    \multirow{1}{*}{SORT} \cite{bewley2016simple}      & YOLOX\cite{yolox2021}                     & \cmark                      & 47.80   & 88.20    & 48.30    \\
    \multirow{1}{*}{DeepSORT} \cite{wojke2017simple}  & YOLOX\cite{yolox2021}                    & \cmark                      & 45.80    & 87.10    & 46.80    \\
    \multirow{1}{*}{MOTDT \cite{Chen2018RealTimeMP}}  & YOLOX\cite{yolox2021}                 & \cmark                      &   39.20   &   84.30   &    39.60  \\
    \multirow{1}{*}{ByteTrack} \cite{zhang2021bytetrack} & YOLOX\cite{yolox2021}                 & \cmark                      & 47.10     & \textbf{88.20}    & 51.90    \\ 
    \multirow{1}{*}{OC-SORT} \cite{cao2023observation} & YOLOX\cite{yolox2021}                 & \cmark                      &  52.10    &   87.30   &    51.60  \\ 
    \bottomrule

    \multirow{1}{*}{$\mathtt{MA-SORT}$}\textbf{(Ours)} & YOLOX\cite{yolox2021}                 & \cmark                      &   \textbf{53.44 }  &  87.31  &  \textbf{53.78  } \\                               
   $\mathtt{MA-SORT}$\textbf{(Ours)}&  $\mathtt{iGLIP}$ $\mathtt{(Z-GMOT)}$\textbf{(Ours)} & \xmark                      &   47.57  & 83.11   &  46.58   \\ \bottomrule

    \end{tabular}
}
    \label{tab:Abl_DanceTrack}
\end{table}

\begin{table}[!t]
\setlength{\tabcolsep}{5pt}
\renewcommand{\arraystretch}{1.0}
  \centering
  \caption{\underline{Ablation study of effectivess} of $\mathtt{MA-SORT}$ on \textit{MOT20} testset with \textit{MOT task}. As ByteTrack, OC-SORT (gray) uses different thresholds for test set sequences and offline interpolation procedure, we also report scores by disabling these as ByteTrack$^\dag$, OC-SORT$^\dag$.  The best scores are highlighted in \textbf{bold}. }
  \vspace{-1em}
  \label{tab:gmot40}
  \resizebox{\columnwidth}{!}{%
\begin{tabular}{l|ccc}
\toprule
\textbf{Trackers}  &  \textbf{HOTA}$\uparrow$ & \textbf{MOTA}$\uparrow$ & \textbf{IDF1}$\uparrow$  \\ \hline
MeMOT \cite{cai2022memot}    & 54.1 & 63.7 & 66.1        \\  
FairMOT \cite{zhang2021fairmot}    & 54.6 & 61.8 & 67.3         \\  
TransTrack \cite{transtrack}   & 48.9 & 65.0 & 59.4         \\  
TrackFormer \cite{meinhardt2021trackformer}    & 54.7 & 68.6 & 65.7   \\ 
ReMOT \cite{yang2021remot}     & 61.2 & 77.4 & 73.1      \\  
GSDT \cite{Wang2020_GSDT}    & 53.6 & 67.1 & 67.5    \\  
CSTrack  \cite{liang2022cstrack}   & 54.0 & 66.6 &  68.6  \\  
TransMOT \cite{chu2023transmot}    & - & 77.4 &  75.2  \\ 
\color{gray}ByteTrack\cite{zhang2021bytetrack}     & \color{gray}61.3 & \color{gray}{77.8}  & \color{gray}75.2  \\  
\color{gray}OC-SORT\cite{cao2023observation}   &  \color{gray}62.4 &	\color{gray}75.7 &	\color{gray}76.3  \\ 
ByteTrack$^\dag$\cite{zhang2021bytetrack}     & 60.4 & 74.2  & 74.5 \\ 
OC-SORT$^\dag$\cite{cao2023observation}   & 60.5 & 73.1  & 74.4 \\ \hline
$\mathtt{MA-SORT}$\textbf{(Ours)} & \textbf{61.4} & \textbf{77.6 }& \textbf{75.5 } \\
\bottomrule

\end{tabular}}
\vspace{-2em}
\label{tb:Abl_MOT20}
\end{table}

\begin{figure*}[t]
\centering
\includegraphics[width=1.0\textwidth]{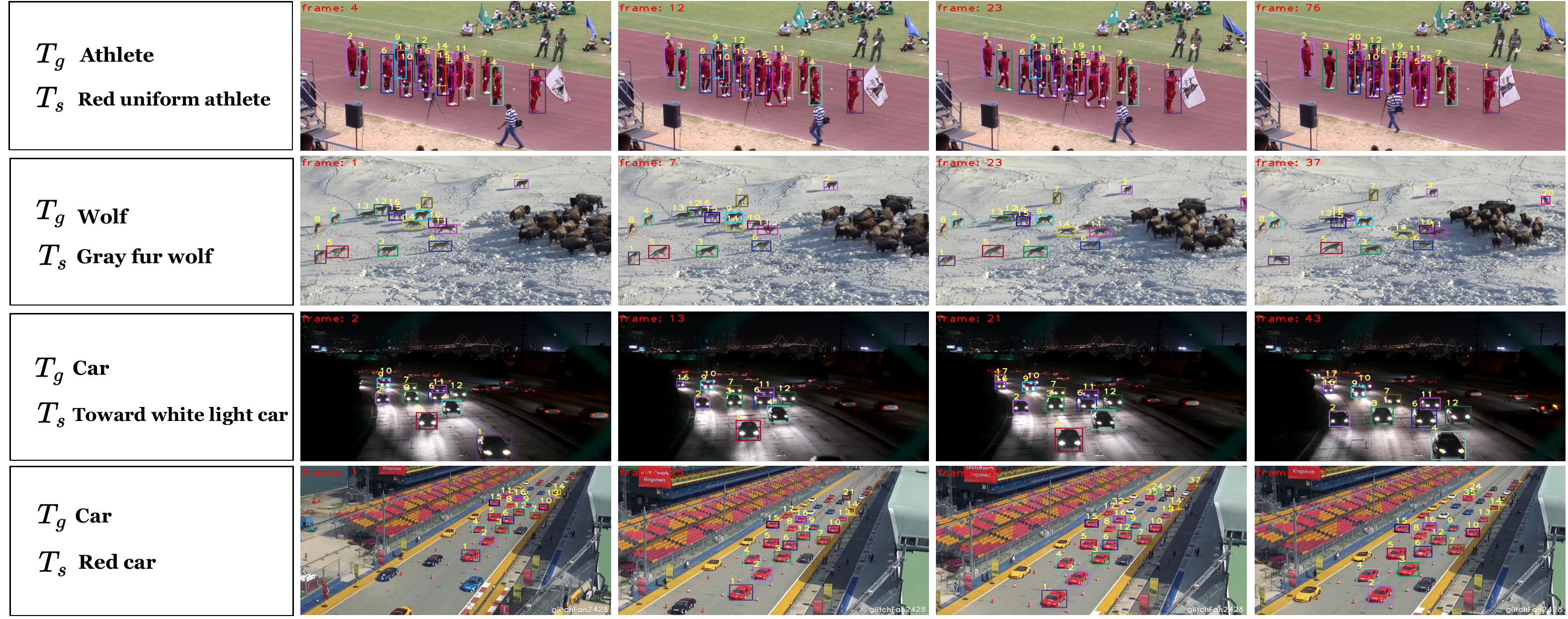}
\vspace{-2em}
\caption{Examples of tracking conducted by our proposed $\mathtt{Z-GMOT}$ using input texture descriptions (left). The texture description including both a general prompt $T_g$, and a specific prompt $T_s$ are integrated into our proposed $\mathtt{Z-GMOT}$ framework including $\mathtt{iGLIP}$ and $\mathtt{MA-SORT}$.}
\vspace{-1em}
\label{fig:iGLIP-uniform}
\end{figure*}


\subsection{Performance Comparison} 

In Table \ref{tb:GMOT40-iGLIP}, we benchmark the tracking performance in two scenarios: one involving the use of one-shot object detection (OS-OD) and the other utilizing our proposed zero-shot $\mathtt{iGLIP}$ on our newly introduced \textit{Refer-GMOT40} dataset.
It is important to note that incorporating OS-OD with these trackers is equivalent to achieving SOTA OS-GMOT \cite{bai2021gmot}. Table \ref{tb:GMOT40-iGLIP} clearly shows that our zero-shot $\mathtt{iGLIP}$, without requiring any prior knowledge or training, achieves significant performance advantages across various metrics when compared to OS-OD, which relies on initial bounding boxes and is run five times. For instance, on OC-SORT tracker, $\mathtt{iGLIP}$ shows improvements in HOTA, MOTA, and IDF1 by 26.16, 42.16, and 33.03 points, respectively. On average across all trackers, $\mathtt{iGLIP}$ outperforms OS-OD by 21.64, 35.67, and 26.61 points in HOTA, MOTA, and IDF1 metrics. 

Table \ref{tb:GMOT40-MASORT} shows the comparison between our proposed $\mathtt{MA-SORT}$ with various trackers using the same object detection, i.e., the proposed $\mathtt{iGLIP}$ on \textit{Refer-GMOT40} dataset. It is evident that $\mathtt{MA-SORT}$ consistently outperforms other trackers. For example, $\mathtt{MA-SORT}$ outperforms DeepSORT and Deep-OCSORT by 6.3, 5.63, 10.62 points and 0.86, 0.6, 1.65 points across all metrics, respectively.

Table \ref{tb:Animal-ZGMOT} presents a comparison between our proposed $\mathtt{Z-GMOT}$ and other existing fully-supervised MOT methods on the \textit{Refer-Animal} dataset. In order to ensure a fair comparison, we have also implemented a fully-supervised $\mathtt{MA-SORT}$ method with YOLOX object detection. While our $\mathtt{MA-SORT}$ with YOLOX object detector achieves the best performance, it is worth noting that $\mathtt{Z-GMOT}$ outperforms other fully-supervised MOT methods without the need for any training data.

\subsection{Ablation Study}

\noindent
\underline{Generalizability of $\mathtt{Z-GMOT}$ framework.} In addition to the GMOT task, we also evaluate its generalizability on the MOT task, as in Table \ref{tab:Abl_DanceTrack} on DanceTrack dataset. This table presents a comparison of $\mathtt{Z-GMOT}$ with the existing fully-supervised MOT methods. To ensure a fair comparison, we have implemented a fully-supervised $\mathtt{MA-SORT}$ method with YOLOX object detection. While our $\mathtt{MA-SORT}$ with YOLOX achieves the best performance, it is noteworthy that $\mathtt{Z-GMOT}$ demonstrates compatibility with SOTA fully-supervised MOT methods, even surpassing SORT, DeepSORT, and MOTDT, all without requiring any training data. In this experiment, both general prompt and specific prompt are set as ``dancer''.

\noindent
\underline{Effectiveness of proposed $\mathtt{MA-SORT}$.} We assess its performance by conducting a comparison on the MOT20 dataset, as outlined in Table \ref{tb:Abl_MOT20}, focusing on the MOT task. To ensure a fair comparison, we disable certain ad-hoc settings that employ varying thresholds for individual sequences and an offline interpolation procedure. In this experiment, we employed the YOLOX object detector, which demonstrates the effectiveness of $\mathtt{MA-SORT}$.

\noindent
\underline{Effectiveness of proposed $\mathtt{iGLIP}$.}
We evaluate our $\mathtt{iGLIP}$ by comparing it to GLIP \cite{li2022grounded} and OS-OD \cite{huang2020globaltrack} for object detection on the \textit{Refer-GMOT40} dataset, as presented in Table \ref{tab:Abl_hyper-param}(a). $\mathtt{iGLIP}$ outperforms other detector methods, achieving the highest scores. It is worth highlighting that despite being an extension of GLIP, $\mathtt{iGLIP}$ exhibits significant improvements, with a 0.7\% increase in $AP_{50}$, a 5.0\% improvement in $AP_{75}$, and a 3.9\% enhancement in $mAP$, demonstrating its clear superiority over GLIP.

\begin{table}[!h]
\centering
    \vspace{-1em}
    \setlength{\tabcolsep}{1pt}
    \centering 
    \caption{Ablation studies on \textit{Refer-GMOT40}.}
    \vspace{-1em}
    \resizebox{1.0\linewidth}{!}{
    \setlength{\tabcolsep}{1pt}
    \subfloat[Object detection by $\mathtt{iGLIP}$. \vspace{-0.5em}]{
    \resizebox{0.5\linewidth}{!}{
    \begin{tabular}{lccc}
    \toprule
    \textbf{Detectors} & $AP_{50}$ & $AP_{75}$  & $mAP$\\ \hline
    OS-OD & 31.5 &13.4 &15.8 \\
    GLIP  & 66.2 & 35.0 &  36.1\\ 
    \hline
    iGLIP &\textbf{66.9}  &\textbf{40.0} &\textbf{40.0}  
      \\ \bottomrule
    \end{tabular}}
    }
    \quad
    \subfloat[Tracking performance with varied $\theta$. \vspace{-0.5em}]{
    \resizebox{0.5\linewidth}{!}{
    \begin{tabular}{l|ccc}
    \toprule
    $\theta$      & \textbf{HOTA}$\uparrow$                  & \textbf{MOTA}$\uparrow$                  & \textbf{IDF1}$\uparrow$                             \\ \hline
    22.5$^\circ$      & 56.57 & 64.57 &  67.85               \\
    45$^\circ$      & 56.58 & 64.59 & 67.89                \\
    67.5$^\circ$  & 56.75 & 64.62 & 68.17 \\
    80$^\circ$ & 56.74 & 64.62 & 68.15 \\ \bottomrule
                              
    \end{tabular}}
    }
    }
    \vspace{-1em}
    \label{tab:Abl_hyper-param}
\end{table}

\begin{table*}[!thb]
\centering
\setlength{\tabcolsep}{5pt}
\renewcommand{\arraystretch}{0.9}
\caption{Properties and computational resources required by our proposed Z-GMOT. Inference time represents an average of 4 videos comprising a total of 1,467 frames.}
\resizebox{.9\linewidth}{!}{
\begin{tabular}{l|cccc|cc}
\toprule
& \multicolumn{4}{c|}{\textbf{ MA-SORT + iGLIP}} & \multicolumn{2}{c}{\textbf{MA-SORT + YOLOX}} \\ \hline
\rowcolor{gray!20}
\multicolumn{7}{l}{\textbf{Properties}} \\ \hline
Settings & \multicolumn{4}{c|}{\textbf{Open-set}} &  \multicolumn{2}{c}{Close-set} \\ \hline
Track Agnotic Objects & \multicolumn{4}{c|}{\textbf{\cmark}} &  \multicolumn{2}{c}{\xmark} \\ \hline
\rowcolor{gray!20}
\multicolumn{7}{l}{\textbf{Computational Cost}} \\ \hline
& Text Encoder & Vision Encoder & DyHead \& RPN & Entire Model  & Vision Encoder & Entire Model \\ \hline
Model Size (\#Params) & 108M & 197M & 122M & 427M & 99.1M & 124.5M \\ \hline
\begin{tabular}[c]{@{}l@{}}Inference time \\ (seconds/frame)\end{tabular} & 0.008 & 0.019 & 0.17 & 0.197 & 0.064 & 0.069 \\ \hline
FLOPs (G) & 45.94 & 181.32 & 136.91 & 364.17 & 281.9 & 322 \\ \hline
GPU memory (Gb) & 1.27 & 2.84 & 2.09 & 6.2 & 8.6  & 10.2 \\ \bottomrule
\end{tabular}}
\label{tb:computation}
\end{table*}

\begin{table*}[!thb]
\centering
\caption{Comparison of tracking performance and computational complexity between RMOT ~\cite{wu2023referring} and our MA-SORT with YOLO-X object detection. We report on 2 classes of human and car because RMOT was trained on only those two classes.}
\resizebox{0.9\linewidth}{!}{
\begin{tabular}{l|lll|lll|cccc}
\toprule
\multirow{3}{*}{\textbf{Methods}} & \multicolumn{6}{c|}{\textbf{Tracking Performance}} & \multicolumn{4}{c}{\textbf{Computational Complexity}} \\ \cline{2-11}
& \multicolumn{3}{c|}{\textbf{Human}} & \multicolumn{3}{c|}{\textbf{Car}} & \multirow{2}{*}{\begin{tabular}[c]{@{}c@{}}Model \\ size\end{tabular}} & \multirow{2}{*}{FLOPs} & \multirow{2}{*}{\begin{tabular}[c]{@{}c@{}}GPUs \\ Usage\end{tabular}} & \multirow{2}{*}{\begin{tabular}[c]{@{}c@{}}Inference \\ Time\end{tabular}} \\ \cline{2-7}
          & HOTA    & MOTA   & IDF1   & HOTA   & MOTA   & IDF1  &            &       &            &                \\ \hline
RMOT~\cite{wu2023referring}    &  1.075       &   -0.55     &   1.19     & 6.57       &  2.99      & 5.41      &  169M          &  212G     &    3 GB        & 0.118 s/f               \\ \hline
MA-SORT + iGLIP &  47.02      & 55.88      &  52.22       & 57.8       & 57.66        &  71.54     &   427M         & 364.17G       &       6.2 GB     &   0.197 s/f    \\ \hline       
MA-SORT + YOLOX  &   33.08      & 39.00       &    41.44    &  29.88      &  22.71      &  34.93       &  124.5M          &  322G     &  10.2  GB        &  0.069 s/f    \\ \bottomrule       
\end{tabular}}
\label{tb:computation-performance}
\end{table*}

\noindent
\underline{Hyper-param $\theta$.}
Table \ref{tab:Abl_hyper-param}(b) shows ablation study of vector distance threshold $\theta$ as defined in Eq.\ref{eq:theta}. The minor variations in tracking performance demonstrate the robustness of our proposed $\mathtt{MA-SORT}$ when $\theta$ is varied within the range of [22.5$^\circ$, 80$^\circ$].
We select $\theta = 67.5$ in the reported results.

\subsection{Computational Complexity}

To evaluate the computation cost, we report the computational resource of each relevant component and inference time as in Table \ref{tb:computation}. It is important to note that the reported inference time represents an average, calculated over 4 videos comprising a total of 1,467 frames. All the implementation and comparison have been conducted on A100 40GB.

In Table \ref{tb:computation}, we report our computational complexity in two scenarios: (i) open-set setting where the proposed iGLIP is used to detect unseen categories. (ii) close-set setting where YOLOX is used to detect pre-defined class. In both scenarios, we use our proposed MA-SORT as an object association.

To evaluate the effectiveness of our proposed Z-GMOT, we suggest to compare with other state-of-the-art methods in the field, focusing on both computational complexity and performance, as detailed in Table \ref{tb:computation-performance}. Included in this comparison is RMOT ~\cite{wu2023referring}, a state-of-the-art model in referring-MOT. It is important to note that RMOT is based on fully-supervised learning and operates within a close-set environment, specifically targeting the tracking of persons and cars. The analysis and comparisons presented in Tables \ref{tb:computation} and \ref{tb:computation-performance} reveal that our Z-GMOT not only holds a comparable computational complexity with state-of-the-art referring tracking methods but also surpasses them with substantial margins.

\section{CONCLUSION \& DISCUSSION}

In this study, we present $\mathtt{Z-GMOT}$, a novel tracking framework capable of tracking diverse objects without relying on labeled data. $\mathtt{Z-GMOT}$ adopts a tracking-by-detection paradigm and offers two key contributions: (i) zero-shot $\mathtt{iGLIP}$ for effective object detection using natural language descriptions and (ii) $\mathtt{MA-SORT}$ for efficient tracking of visually similar objects within a broader context of generic objects. Beyond proposing $\mathtt{Z-GMOT}$, we also introduce a new \textit{Refering GMOT dataset}. We have thoroughly assessed and demonstrated the efficacy and adaptability of $\mathtt{Z-GMOT}$, not only in the GMOT task but also in the MOT task.

\noindent \textbf{Discussion.} We utilize GLIP as our preferred VLM for developing $\mathtt{iGLIP}$. However, it is important to recognize the rich diversity of VLMs available in the field, which opens up exciting avenues for deeper exploration. Moreover, in our current study, we have implemented $\mathtt{Z-GMOT}$ exclusively using only textual description \texttt{object} and \texttt{attributes}. Nevertheless, our Referring GMOT dataset offers additional information, such as \texttt{object\_synonyms} and \texttt{other\_attributes}, which hold great potential for further research, particularly in the context of prompt tuning or prompt engineering. Exploring these additional aspects of our Referring GMOT dataset could lead to enhanced object tracking capabilities as well as other fields such as surveillance, robotics, and animal welfare. We expect our work to inspire future research in the unexplored realm of unseen MOT/GMOT paradigms, potentially leading to extensions in other tracking scenarios, e.g., open-vocabulary MOT/GMOT.

\newpage

\footnotesize
\newlength{\bibitemsep}\setlength{\bibitemsep}{.6\baselineskip plus .05\baselineskip minus .05\baselineskip}
\newlength{\bibparskip}\setlength{\bibparskip}{0pt}
\let\oldthebibliography\thebibliography
\renewcommand\thebibliography[1]{%
  \oldthebibliography{#1}%
  \setlength{\parskip}{\bibitemsep}%
  \setlength{\itemsep}{\bibparskip}%
}
\bibliography{main}
\bibliographystyle{acl_natbib}




\end{document}